# A review of clustering models in educational data science towards fairness-aware learning

**Tai Le Quy**[1], **Gunnar Friege**[2], **Eirini Ntoutsi**[3]

[1]L3S Research Center, Leibniz University Hannover
Address: Appelstr. 4, 30167 Hannover, Germany
Email: tai@l3s.de
ORCID: 0000-0001-8512- 5854

[2]Institute for Didactics of Mathematics and Physics, Leibniz University Hannover
Address: Welfengarten 1A, 30167 Hannover, Germany
Email: friege@idmp.uni-hannover.de
ORCID: 0000-0003-3878-9230

[3]Research Institute CODE, Bundeswehr University Munich
Address: Carl-Wery-Str. 18, D-81739 München, Germany
Email: eirini.ntoutsi@unibw.de
ORCID: 0000-0001-5729-1003

**Abstract**. Ensuring fairness is essential for every education system. Machine learning is increasingly supporting the education system and educational data science (EDS) domain, from decision support to educational activities and learning analytics. However, the machine learning-based decisions can be biased because the algorithms may generate the results based on students' protected attributes such as race or gender. Clustering is an important machine learning technique to explore student data in order to support the decision-maker, as well as support educational activities, such as group assignments. Therefore, ensuring high-quality clustering models along with satisfying fairness constraints are important requirements. This chapter comprehensively surveys clustering models and their fairness in EDS. We especially focus on investigating the fair clustering models applied in educational activities. It is believed that these models are practical tools for analyzing students' data and ensuring fairness in EDS.

## Abbreviations

| | |
|---|---|
| ACC | Clustering accuracy |
| AI | Artificial Intelligence |
| AIED | Artificial Intelligence in Education |
| ANOVA | Analysis Of Variance |

| ARI | Adjusted Rand Index |
| BIRCH | Balanced Iterative Reducing and Clustering using Hierarchies |
| BMI | Body Mass Index |
| BMU | Best Matching Unit |
| CFSFDP-HD | Clustering by Fast Search and Finding of Density Peaks via Heat Diffusion |
| CHI | Calinski–Harabasz index |
| CLARA | Clustering in LARge Applications |
| CLARANS | Clustering Large Applications based on RANdomized Search |
| CORE | Computing Research and Education Association of Australasia |
| DBSCAN | Density-based spatial clustering of applications with noise |
| DBI | Davies–Bouldin Index |
| DBLP | Database Systems and Logic Programming |
| DI | Dunn Index |
| DP | Dirichlet process |
| EDM | Educational Data Mining |
| EDS | Educational Data Science |
| EM | Expectation–Maximization |
| EMT | Ensemble Meta-based Tree |
| FCM | Fuzzy c-means clustering |
| FIE | Frontiers in Education |
| ICALT | International Conference on Advanced Learning Technologies |
| ITS | Intelligent Tutoring Systems |
| KPCA | Kernel-based Principal Component Analysis |
| LA | Learning Analytics |
| LD | Learning Design |
| LMS | Learning Management System |
| ML | Machine Learning |
| MIT | Massachusetts Institute of Technology |
| MOOC | Massive Open Online Course |
| NMI | Normalized mutual information |
| OPTICS | Ordering Points To Identify the Clustering Structure |
| OULAD | Open University Learning Analytics |
| PAM | Partition around medoids |
| PISA | Program for International Student Assessment |
| RQ | Research question |
| SJR | SCImago Journal & Country Rank |
| SOM | Self-organizing maps |
| SSE | Sum of Squared Error |
| SVM | Support Vector Machine |

# 1 Introduction

Fairness is a fundamental concept of education, whereby all students must have an equal opportunity in study or be treated fairly regardless of, e.g., their household income, assets, gender, race, or knowledge and domain-specific abilities. Fairness in the education system is reflected in a wide range of education-related activities, such as assessment and measurement (Dorans & Cook, 2016; Zlatkin-Troitschanskaia et al., 2019), students' group work and group assignment (Rezaeinia, Góez, & Guajardo, 2021; Miles & Klein, 1998; Ford & Morice, 2003), graduate school admission (Song, 2018), predicting student performance (Xiao, Ji, & Hu, 2021). One of the kernel demands for justice is education; therefore, having a fair education system is crucial to achieving justice in society (Meyer, 2014).

In EDS, machine learning (ML) has been used in a wide variety of decision-making and learning analytics (LA) and educational data mining tasks (Peña-Ayala, 2014; Dutt, Ismail, & Herawan, 2017; Romero & Ventura, 2020; McFarland, Khanna, Domingue, & Pardos, 2021); for example, student dropout prediction (Del Bonifro, Gabbrielli, Lisanti, & Zingaro, 2020; Kemper, Vorhoff, & Wigger, 2020), education admission decisions (Song, 2018) or forecasting on-time graduation of students (Livieris, Tampakas, Karacapilidis, & Pintelas, 2019; Hutt, Gardner, Duckworth, & D'Mello, 2019). The results of these ML models are the basis for building applications in EDS, such as student data analysis, learning support, and decision support systems. In fact, different ML-based decisions can be made based on protected attributes (i.e., the attributes for which the model is likely to exhibit bias), such as gender or race, leading to discrimination (Ntoutsi et al., 2020). Hence, improving fairness w.r.t. protected attributes in the results of ML models is imperative while maintaining the performance of the models. In other words, ensuring fairness in ML models also contributes to equity in educational systems.

Research on fairness-aware ML has been carried out in various domains such as finance, education, healthcare, criminology and social issues (Mehrabi, Morstatter, Saxena, Lerman, & Galstyan, 2021; Le Quy et al., 2022). However, along with the development of the EDS research community, there are more and more studies on ensuring the fairness of ML models applied in education. These studies mainly focus on supervised learning models on the students' data (Gardner, Brooks, & Baker, 2019; Riazy, Simbeck, & Schreck, 2020; Bayer, Hlosta, & Fernandez, 2021). Recently, there have been several surveys on algorithmic bias and fairness in education, in which the authors presented the theory of fairness and summarized the classification and predictive methods (Baker & Hawn, 2021; Kizilcec & Lee, 2022).

Clustering is a novel and important technique in EDS (McFarland et al., 2021) and educational data mining (Dutt et al., 2017). The main operation of clustering models is to cluster instances, i.e., students, into groups based on similarity measures to discover unknown patterns in the educational data, such as under-standing student achievement (S. Liu & d'Aquin, 2017), characterizing students' learning behaviors (N. Zhang et al., 2017). However, there are few studies on fairness in clustering models in education (Le Quy et al., 2021). Furthermore, an overview of clustering models and their fairness applied in education is still missing. Hence, it is necessary to conduct a survey of the commonly used clustering models in EDS problems and discuss the fairness of these clustering models. Therefore, our survey is serving to fill such a gap in the extant research. We believe that our survey is useful as it explores and consolidates the technical details of clustering algorithms and fairness constraints on the educational data. Hence, it will help both educational scientists and computer scientists easily select the most appropriate methods for their clustering tasks that require fairness, such as student assessment and group assignment, and to promote this subfield of research within EDS.

The review is organized around the following key research questions:

- $RQ_1$. Why is clustering an important task for EDS? What clustering models and for what purposes have been applied in EDS?
- $RQ_2$. What is fairness in clustering? How to achieve fair clustering results? How to define and achieve fairness in clustering in EDS?
- $RQ_3$. How to evaluate the quality of (fair) clustering in EDS? How is experimental evaluation (datasets, parameter setting, methods, quality measures, etc.) performed? Are there benchmark datasets?
- $RQ_4$. Beyond fairness, what are other issues/requirements to be considered for clustering in EDS?

In the following subchapters, we will summarize the EDS topics/problems where the clustering models have been used. We will present the most prevalent used clustering models w.r.t. their abilities in Section 2. Next, in Section 3, we will overview existing fair clustering models and consider their potential for EDS. After that, we will discuss evaluation aspects from benchmark datasets for (fair) clustering in EDS to evaluation measures in Section 4. Requirements for clustering and other aspects in EDS beyond fairness will be covered in Section 5. Finally, Section 6 will conclude this survey with an outlook for future work.

## 2 Clustering models for EDS

In this section, we answer the research question $RQ_1$ regarding the role of clustering in EDS and its typical applications. We also overview the main clustering methods used in EDS and discuss several aspects of practical importance beyond the algorithm itself, like cluster model assumptions and complexity.

## 2.1 Methodology of the survey process

We perform the following procedure to address *RQ₁*. First, we search the relevant works by a set of keywords in three well-known scientific databases (Google Scholar[1], DBLP[2] and Scopus[3]) and select only quality studies on ranked venues; second, we categorize and summarize the educational tasks related to clustering in Sections 2.2 and outline the different clustering methods in Section 2.3.

In the first step, we identify the following relevant keywords: *bias, fairness, clustering, learning analytics, educational data mining,* and *educational datasets*. We define a set of synonyms and antonyms for each, as listed in Table 1. The search queries are performed on Google Scholar and DBLP with all possible combinations of the different synonyms (antonyms) of the keywords. After identifying the papers, we select only the published papers on ranked conferences (A*, A, B, C) or journals (Q1, Q2, Q3, Q4) that are indexed by Scopus. We refer to the updated databases, i.e., issued in 2021, of Computing Research and Education Association of Australasia (CORE)[4] for conferences' ranking and SCImago Journal & Country Rank (SJR)[5] for journals' ranking. Because the related research of Dutt et al. (2017) provided a systematic literature review on clustering algorithms and their applicability and usability in EDM from 1983 to 2016, we consider 133 publications from 2017 to June 2022.

In the next step, we analyze metadata on the selected publications. Table 2 provides a brief regarding the number of publications on venues w.r.t. ranking. As shown in Table 2, 29.3% (39 out of 133) of selected articles are published in leading conferences/journals, i.e., Q1/A*/A ranking. The papers are collected from 97 venues (47 conferences and 50 journals). The top 10 venues are listed in Table 3; they account for 31.6% (42 out of 133) of the total selected papers. In addition, we count the number of publications by year and visualize it in Fig. 1. There has been a slight upward trend in the number of articles published in peer-reviewed journals over the years.

---

[1] https://scholar.google.com/

[2] https://dblp.org/

[3] https://www.scopus.com/

[4] http://portal.core.edu.au/conf-ranks/

[5] https://www.scimagojr.com/

**Table 1** A set of keywords

| Keywords | Synonyms | Antonyms |
|---|---|---|
| bias | discrimination | fair, unbiased |
| fairness | equity, justice | |
| clustering | grouping | |
| learning analytics | | |
| educational data mining | | |
| educational datasets | | |

**Table 2** Venue-based distributions of publications

| Type of venues | Q1/A/A* | Q2/B | Q3/C | Q4/Other |
|---|---|---|---|---|
| Journal | 25 | 18 | 7 | 13 |
| Conference | 14 | 21 | 5 | 30 |

**Table 3** The top 10 venues

| No | Venues | #Publications | Ranking |
|---|---|---|---|
| 1 | Educational Data Mining (EDM) | 11 | B |
| 2 | Journal of Physics: Conference Series | 7 | Q4 |
| 3 | International Conference on Artificial Intelligence in Education (AIED) | 5 | A |
| 4 | IEEE Access | 4 | Q1 |
| 5 | Education and Information Technologies | 3 | Q1 |
| 6 | Scientific Programming | 3 | Q2 |
| 7 | IEEE Frontiers in Education Conference (FIE) | 3 | C |
| 8 | Complexity | 2 | Q1 |
| 9 | International Conference on Advanced Learning Technologies (ICALT) | 2 | B |
| 10 | International Conference on Intelligent Tutoring Systems (ITS) | 2 | B |

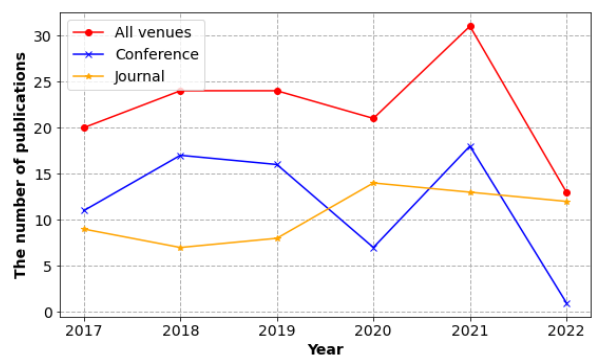

**Fig. 1** Number of publications over the years (2022 is not yet complete).

## 2.2 EDS tasks using clustering models

Clustering has been used in a variety of tasks in EDS, from analyzing students' behavior (Section 2.2.1) and performance (Section 2.2.2) to grade prediction (Section 2.2.3), recommendations (Section 2.2.4), supporting collaboration and teamwork (Section 2.2.5) and analyzing students' wellbeing (Section 2.2.6). Student data are collected from various sources, including traditional classrooms and learning management systems (LMS), such as Moodle[6] or ITS.

### 2.2.1 Analyzing students' behavior, interaction, engagement, motivation and emotion

Clustering methods are used to discover the relationship between students' behaviors and their learning performance (Fang et al., 2018; W. Chang et al., 2020; S. Zhang et al., 2021; Varela et al., 2019; Mai et al., 2022). This allows teachers to get a good overview of their students and manage them accordingly (Ding et al., 2017). For example, Waspada et al. (2019) used $k$-means to cluster students' behavior and discovered that the activity of working on online quizzes affects the final scores. Students who participated more became more engaged in the course and achieved higher grades (Esnashari, Gardner, & Watters, 2018). Jia et al. (2017) clustered the behavioral records of reading tests in an online Chinese reading assessment system to identify students' reading abilities and testing strategies. Howlin & Dziuban (2019) proposed a repeated fuzzy clustering algorithm for discovering student behaviors or outliers. Students' behavior might change over time and therefore, it is necessary to investigate such changes and ensure an up-to-date clustering model. McBroom et al. (2020) applied a hierarchical clustering method to detect the behavioral trends of students over time and produced clusters of students' behavior, which characterize exercises where many students give up. Shen and Chi (2017) developed a temporal clustering framework to determine if a student's learning experience is "unprofitable" to offer a personal suggestion.

---

[6] https://moodle.org/

Since gamification is starting to play an increasingly important role in education activities, Ruipérez-Valiente et al. (2017) analyzed the relationship between students' interactions and behaviors with the badge system, i.e., students earn badges through their interaction with engineering problems and the learning process. The results are considered as feedback or guidelines on how to provide appropriate game-based tasks to students, which can help them improve their learning. The interactions of university students in an e-learning system are clustered and analyzed by the hierarchical clustering algorithm and $k$-means algorithm in order to discover the relationship between the final grades and the use of the modules (López et al., 2017). Besides, experiments were performed on several clustering methods to cluster students' interactions and investigate the associations between students' academic performance and social interactions (Mengoni et al., 2018). The resulting clusters can help educators identify the special students and improve their Learning Design (LD) process and the teaching materials that need to be revised.

Using the expectation–maximization (EM) clustering algorithm, Orji and Vassileva (2020) discovered that students' engagement is a good indicator of academic performance. Therefore, categorizing and identifying students with low levels of engagement are effective ways to help them improve their academic performance (Khalil & Ebner, 2017; Roy et al., 2017; Güvenç & Çetin, 2018; Oladipupo & Olugbara, 2019; Palani et al., 2021). Besides, different levels of engagement are identified in students who use different learning strategies through Pearson correlation analysis on clusters resulting from the agglomerative hierarchical clustering method (J. B. Huang et al., 2021). $k$-means algorithm was used to cluster students based on 12 engagement metrics (Moubayed et al., 2020). The authors discovered that the most representative of the students' engagement levels are the number of logins and the average duration to submit assignments.

Students' motivation is essential in the massive open online course (MOOC) environment (Hartnett, 2016). $k$-means was used to analyze the questionnaire results of the MOOC system (Nen-Fu et al., 2018) and discover the relationship between learning motivation and the learning behavior of the students. Besides, self-motivation in college students was investigated by using $k$-means clustering to reveal the effect of a hidden curriculum and character-building variables. The experiments were performed on the real dataset collected at the State University of Malang, Indonesia (Gunawan et al., 2018). In addition, teachers should also pay attention to students' emotions (Z. Wang & Wang, 2022) because there is a relationship between emotion and performance (Ashkanasy, 2004) as well as between the emotions and motivation of the students (Muñoz-Merino et al., 2014). Students can obtain many emotional experiences through physical education activities, which improves learning interest and relationships among teachers and students (H. Guo & Wang, 2022).

### 2.2.2 Analyzing students' performance and grading

Learning analytics (LA) is an important research area of EDS, which focuses on the collection and analysis of data about learners. In LA, students are often divided into groups based on their performance, which allows the educators to identify low-performing students and help them accordingly (Salwana et al., 2017; Vo & Nguyen, 2018; Thilagaraj & Sengottaiyan, 2020), whereas for high-performing students, enrichment tasks can be assigned to help them advance further. Moreover, analyzing student learning outcomes is one of the most effective methods for understanding the factors affecting students' learning ability (Bharara et al., 2018; Yin, 2021). For example, procrastination is an important indicator of students' performance, with non-procrastinators tending to have higher performance (Park et al., 2018; Hooshyar et al., 2019). Using $k$-means, experimental results show that the academic performance of female students is affected by nutrition and health issues (Preetha, 2021). In this direction, $k$-means clustering model was used to discover the factors contributing to students' performance Aghababyan et al. (2018), resulting in interpretable clusters of students based on their confidence entropy[7], degree of over/under confidence and other related variables.

A clustering model not only can help teachers analyze students' assignments, but it is also an effective method to support essay grading. For instance, interpretable clustering has been used to cluster and find the structure of students' solutions in a Python programming course (Effenberger & Pelánek, 2021). With a similar purpose to automatically cluster programming assignments (C programming), $k$-means was applied in the study of Gao et al. (2019). Besides, L.-H. Chang et al. (2021) proposed a clustering method to cluster sentences to support grading the essays written in Finnish by bachelor students. In addition, Sobral and de Oliveira (2021) used $k$-means to observe the change in self-assessment skills between two evaluation moments (after submitting their work/test and after knowing the correct answers) to investigate the role of self-evaluation in the student's final grade.

### 2.2.3 Predicting students' performance

Student performance prediction (including test scores, students at-risk, dropout, etc. prediction) is a common task in EDS (Khan & Ghosh, 2021). Various EDS techniques, such as correlation, regression, and classification, have been applied to predict students' performance (Romero & Ventura, 2017). By using clustering models, studies have tried to achieve various objectives, which can be grouped broadly into three categories: 1) Predicting the performance of students in terms of the actual grade as a regression problem; 2) Considering student performance prediction as a classification problem; 3) Predicting student at-risk or dropout.

---

[7] Students' confidence entropy is computed by the Shannon equation.

To predict students' grades, clustering has been used as a pre-processing step to detect groups/clusters of similarly performing students before applying the predictive models. For example, $k$-means was used to reveal the unique types of students labeled as "proficient", "struggling", "learning", and "gaming" (Adjei et al., 2017), which was helpful in predicting standardized test scores. Besides, Bayesian fuzzy clustering was used to cluster students into groups (Ramanathan et al., 2019), and then the Kernel-based principal component analysis (KPCA) was applied to identify the best features from the data, which is used in the prediction phase with a Lion-Wolf deep belief network model. Recently, Hassan et al. (2022) applied $k$-means to group students in terms of both static attributes like *sleep, stress, study spaces* and spatio-temporal attributes like *latitude, longitude, time* before transferring these clusters to several well-known regression models for grade prediction.

Casalino et al. (2019) proposed an adaptive fuzzy clustering algorithm to process the educational data as data streams to predict student outcomes (pass, fail). They cluster non-overlapping chunks of data and then investigate cluster prototypes to classify students into two classes (pass, fail). In addition, an Ensemble Meta-based Tree (EMT) model was introduced to classify students into four categories (excellent, very good, good, and satisfactory) (Almasri et al., 2020). In their architecture, $k$-means was used to find a set of homogeneous clusters before applying a classifier model for each cluster. Concerning predicting student enrollment in postgraduate studies, $k$-means was applied to reveal the presence of three coherent clusters of students (Iatrellis et al., 2021). In another approach, Francis and Babu (2019) classified students into three groups (high, medium, low) by using four popular classifiers (SVM, Naive Bayes, Decision Tree, and Neural network). After that, $k$-mean clustering and majority voting are used to predict the best accuracy of students. In the MOOC environment, Y.-W. Chu et al. (2021) used clustering-guided meta-learning-based to exploit clusters of frequent patterns in the students' click-stream sequences in order to predict students' in-video quiz performance.

Many studies predict student dropout and at-risk by applying clustering methods because student dropout is the primary concern of many educational institutes (Khan & Ghosh, 2021), especially in MOOCs. A soft subspace clustering algorithm has been applied to discover the interesting patterns and relationships with student dropout rates (Iam-On & Boongoen, 2017a). A link-based cluster ensemble was proposed to transform the data into a new form before applying some classification models (Iam-On & Boongoen, 2017b) to improve the performance of student dropout prediction. $k$-means was used to discover the students likely to drop out of school (Purba et al., 2018). Regarding student at-risk prediction, a time-series clustering (Hung et al., 2017) was applied to capture the at-risk students. Besides, a clustering ensemble method on temporal educational data combining dynamic topic modeling and kernel $k$-means was introduced to predict early students in-trouble (Nguyen & Vo, 2019).

### 2.2.4 Supporting learning, providing feedback and recommendation

Hierarchical clustering was applied to group students with similar learning styles in the classroom to support teachers in understanding the cognitive skills of students and selecting appropriate teaching methods for each group(Yotaman et al., 2020). Besides, grouping students based on their knowledge patterns is a common strategy to allow lecturers to monitor students' performance easily and provide suitable learning materials to each group (Khayi & Rus, 2019; Qoiriah et al., 2020; Silva & Silla, 2020). In addition, in the tutoring process, the appropriate assignment of students - teachers contributes to improving students' learning quality. Hence, (Urbina Nájera et al. (2017) proposed a clustering model based on $k$-means to cluster students and teachers according to their skills and affinities. Furthermore, an online annotation and clustering system has been developed to allow lecturers to generate students' online reading activity patterns and review their annotations (M.-H. Chang et al., 2017).

In terms of generating automatic feedback for students answering a well-being survey, Kylvaja et al. (2019) applied hierarchical clustering to identify the typical patterns in the well-being profiles. The automatic feedback was generated by searching the best matching cluster for an individual survey response. With the aim of providing feedback on psychological fitness, Y. Li and Sun (2022) applied the fuzzy $c$-means algorithm to analyze the university students' psychological fitness data by extracting the characteristics of students' rebellious psychological phenomenon, e.g., talking during class and continually using profanity. Besides, in programming education, researchers suggested that feedback on students' assignments can be generated based on program repair techniques (Gulwani et al., 2018). The authors proposed an automated program repair algorithm for a fundamental programming course. They built a clustering algorithm to divide similar correct solutions into a group and applied the existing correct solutions to repair the new incorrect assignments. In addition, concerning students' feedback for academic courses, Masala et al. (2021) applied $k$-means on local contexts (where the keywords occurred) centered on specific keywords to find similar students' opinions.

As the requirement of developing a course recommendation method to satisfy students' personal goals and interests in MOOC, Y. Guo et al. (2022) proposed a group-oriented course recommendation approach. They used a semantic $k$-means clustering algorithm to group students before applying a course recommendation algorithm based on similarity and expert knowledge to generate the final recommended courses. Following another approach, hierarchical clustering was used to divide students into different types, and a collaborative filtering AI algorithm was utilized for lesson recommendations (M. Wang & Lv, 2022). Furthermore, H. Liu et al. (2019) introduced an incremental tensor-based adaptive clustering method based on correlative analysis and personalized recommendation to offer students appropriate learning resources in various contexts.

### 2.2.5 Supporting collaboration activities

Teamwork is a popular activity in educational settings and is an essential factor in improving students' engagement in the classroom (Fasanya & Fathizadeh, 2019). Clustering methods have been applied to group learners in traditional classrooms and MOOC systems. In the traditional classroom, students are grouped into homogeneous and heterogeneous groups based on their knowledge levels to capture rich semantic information about the group (Wu et al., 2021). They apply a spectral clustering algorithm for dividing students into groups with a group size of no more than eight. Besides, Pratiwi et al. (2017) proposed their clustering method to automatically generate heterogeneous groups based on students' dissimilarity. The resulting clusters are comparable with the teachers' manual grouping solutions. In addition, $k$-means is applied to produce the training groups for American football student-athletes based on their roles and expected performance during the competition (Shelly et al., 2020).

In MOOC systems, the problem of grouping students w.r.t. preferences and interests was introduced by (Akbar et al., 2018). They introduced a grouping method based on hierarchical $k$-means clustering and a weighting formula (the priority of topic preferences) to satisfy students' interests and to ensure that students are grouped in a team with similar preferences. Furthermore, Y. Wang and Wang (2022) developed their clustering technique based on the enhanced particle swarm optimization algorithm to group students w.r.t. their knowledge state and interests.

### 2.2.6 Analyzing physical and mental health

Health and well-being are determined by an individual's physical condition and the potential to contribute to society (Yang, 2021). In their study, $k$-means was used to cluster and analyze the physical quality of college students based on their characteristics (body mass index, vital capacity, endurance quality, flexibility strength quality, and speed dexterity quality). From there, colleges and universities could optimize their education and teaching plans and perfect their talent training programs by using the study's results. In other aspects, a variety of clustering methods ($k$-means, hierarchical clustering, BIRCH, etc.) were applied to identify several types of students based on fitness scores and BMI (fit, not-fit, overweight-and-fit, normal-weight-and-non-fit) (Dovgan et al., 2019). Besides, since nutrition may affect both students' physical health and well-being, an exploratory analysis was performed to understand university students' foot habits by using $k$-means (Natilli et al., 2018).

In terms of mental health analysis, psychological fitness was analyzed by using fuzzy $c$-means (Y. Li & Sun, 2022). Besides, a fast-clustering method was used to analyze college students' mental health data (Y. Chu & Yin, 2021). As a result, the findings from these studies may support school counselors and student managers in providing better mental health services for students. In addition, Y. Li et al. (2021) explained the relationship between mental health and career planning and examined how mental health education and career planning are integrated for college students using fuzzy feature clustering.

### 2.2.7 Miscellaneous tasks

Differentiated institutions and diversity are key policy issues in higher education (C. Wang & Zha, 2018). Hence, C. Wang and Zha (2018) introduced three different methods to measure diversity in higher education: 1) using existing institutional classifications to measure systemic diversity; 2) measuring systemic diversity based on detecting optimal types of institutions ($k$-means and k-medians were applied to obtain the institutional types from the higher education institutions data); 3) taking into consideration both within-group homogeneity and between-group heterogeneity when measuring systemic diversity. Experimental results based on data from Chinese universities in 1998 and 2011 showed that the operations and profiles of Chinese universities have become more diverse.

The common assumption of similarity-based clustering methods is that students should perform similarly on items that belong to the same knowledge component (Nazaretsky et al., 2019). Therefore, the authors proposed a new item-similarity measure, namely Kappa Learning. As students progress through the items, their mastery of the skills underlying each item changes, and this measure indicates similarity between items. The experimental results on the K-6 Math intelligent tutoring system showed that the new measure outperforms clustering based on popular similarity measures. In clustering, feature selection plays an essential role because it strongly affects the clustering quality (L. Huang et al., 2019). They developed an efficient algorithm to select the most important features using an entropy-based feature selection method. Their proposed method was evaluated on an online course at Guangxi Radio & TV University and showed its effectiveness in choosing good quality features and contribution to the interpretability of the resulting clusters.

Furthermore, concerning implementing an intelligent educational administration system, F. Liu (2021) introduced a student achievement prediction model based on fuzzy *c*-means and collaborative filtering. The problem of grouping leadership variables in non-formal education has been investigated in the research of Rahmat (2017) using hierarchical clustering. Towards the expansion of personalization in online learning environments, Ahmed et al. (2021) used *k*-means to cluster students into similar groups based on their skills. In addition, grouping students can also be based on their learning style (Pamungkas et al., 2021). The DBSCAN algorithm can be used to analyze the learning status of students, which influences teaching activities (Du et al., 2021).

### *2.3 Clustering models*

We identify the clustering models in the literature and record publications using the clustering algorithms in Table 8 (in Appendix). There are 23 clustering methods applied for EDS tasks. *k*- means is the most prevalent approach, followed by hierarchical and fuzzy *c*-means clustering. The distribution of the clustering algorithms across observed publications is illustrated in Fig. 2. Based on the popularity of the methods, we overview the algorithms used in at least three publications.

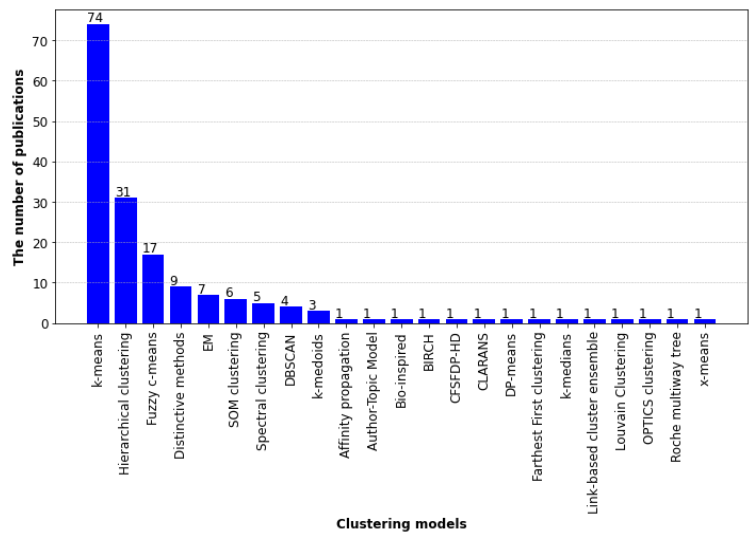

**Fig. 2** Clustering models applied in EDS

The goal of a clustering model is to group objects, e.g., students, into clusters where the objects in the same cluster are similar and the objects in different clus-ters are different. We denote $\mathcal{D} = \{x_1, x_2, \dots, x_n\}$ be a dataset of $n$ objects[8] in $d$-dimensional space, and $k$ is the number of clusters. We list the symbols used in our review in Table 4. In the following sections, we present eight popular clustering models used in EDS.

**Table 4** Symbols and their descriptions

| Symbol | Description |
|---|---|
| $\mathcal{D}$ | Dataset $\mathcal{D} = \{x_1, x_2, \dots\}$ |
| $\mathcal{C}$ | A clustering $\mathcal{C} = \{C_1, C_2, \dots, C_k\}$ |
| $n$ | The number of data points |
| $d$ | The number of attributes (dimensions) |
| $k$ | The number of cluster centers (centroids, medoids) |
| $S$ | A set of cluster centers $S = \{s_1, s_2, \dots, s_k\}$ |
| $x$ or $x_i$ | A data point |
| $C_j$ | A cluster |
| $s$ or $s_j$ | A cluster center (centroid, medoid) |
| $t$ | The number of iterations |
| $dist$ (,) | A distance function |
| $\mathcal{L}(\mathcal{C}, \mathcal{D})$ | An objective function of clustering $\mathcal{C}$ on dataset $\mathcal{D}$ |

### 2.3.1 *k*-means

$k$-means (MacQueen, 1965) is the most popular partitioning-based clustering method that aims to partition $n$ objects into $k$ clusters. The number of clusters $k$ is given in advance. A clustering $\mathcal{C}$ is a partition of dataset $\mathcal{D}$ into $k$ disjoint subsets, $\mathcal{C} = \{C_1, C_2, \dots, C_k\}$, called clusters with $S = \{s_1, s_2, \dots, s_k\}$ be the corresponding cluster centers (centroids). The centroid of each cluster is computed by the mean of the cluster's members. Formally, the goal is to minimize the clustering cost:

$$\mathcal{L}(\mathcal{C}, \mathcal{D}) = \sum_{i=1}^{k} \sum_{x \in C_i} dist(x, s_i)^2. \qquad (1)$$

where function $dist(x, s_i)$ returns the distance from any point $x \in C_i$ to the centroid $s_i$ of cluster $C_i$.

**Algorithm**: Lloyds's version (Lloyd, 1982) is the most prevalent implementation of $k$-mean algorithm, described below:

1. Randomly choose $k$ points as initial centroids $\{s_1, s_2, \dots, s_k\}$.
2. Assign each point in $\mathcal{D}$ to its closest centroid.
3. Update the center of each cluster based on the new point assignments.

---

[8] Because the objects or data points for clustering in the related work are mainly students, we use the terms "objects", "data points" and "students" interchangeably.

4. Repeat until convergence.

In this chapter, there are several variant versions of $k$-means used in the related work, including $k$-medians (Jain & Dubes, 1988) (the median of points in a cluster is used to compute its centroid) and $x$-means (replicate partitions and separates the best results until a certain threshold is reached) (Pelleg et al., 2000).

**Complexity**: The computational complexity of $k$-means is $\mathcal{O}(tkn)$, where $t$ is the number of iterations needed until convergence, and usually $k, t \ll n$.

### 2.3.2 $k$-medoids

$k$-medoids clustering is introduced by Kaufman and Rousseeuw (1990) with the partition around medoids (PAM) algorithm. Similar to $k$-means, the goal of $k$-medoids is to minimize the clustering cost (Eq.1). However, $k$-medoids select the actual data points as cluster centers (namely medoids).

**Algorithm**: The PAM algorithm uses a greedy search strategy, which is presented below:

1. Arbitrarily select $k$ data points as the medoids.
2. Assign each data point to the closest medoid.
3. While the clustering cost decreases:
    a. For each medoid $s$, and for each non-medoid $x$:
        i. Check whether $x$ could replace $s$, and compute the cost change.
        ii. If the cost change is the current best, remember $(s, x)$ combination.
    b. Perform the best $swap(s_{best}, x_{best})$ if it decreases the clustering cost, else terminate the algorithm.

**Complexity**: The computation complexity of the original PAM algorithm per iteration (step 3) is $\mathcal{O}(k(n-k)^2)$. The complexity can be reduced to $\mathcal{O}(n^2)$ with several improved algorithms (CLARA, CLARANS) (Schubert & Rousseeuw, 2021). In which, the CLARANS algorithm was evaluated in the study of Vasuki and Revathy (2022) to predict the degree of student achievement in placement.

### 2.3.3 Hierarchical clustering

Hierarchical clustering is a set of nested clusters organized as a hierarchical tree that can be visualized as a dendrogram (Patel et al., 2015). There are two main types: Agglomerative (bottom-up approach) and Divisive (top-down approach). Hierarchical agglomerative clustering is applied in the related work. BIRCH (T. Zhang et al., 1996) - a version of hierarchical agglomerative clustering which is suitable for very large databases, is used by Dovgan et al. (2019) for their analysis. This section presents the basic hierarchical agglomerative clustering method (Kaufman & Rousseeuw, 1990). There are many approaches for computing the proximity of two clusters (linkage algorithms): single linkage, complete linkage, average linkage, centroid-distance, Ward's method (Ward Jr, 1963), etc.

**Algorithm:**
1. Compute the proximity matrix.
2. Each data point is considered as a cluster.
3. *Repeat*
    a. Merge two closest clusters.
    b. Update the proximity matrix.
*Until* only a single cluster

**Complexity**: The time complexity of the basic agglomerative algorithm is $\mathcal{O}(n^3)$. The complexity can be reduced to $\mathcal{O}(n^2 \log n)$ by using the heap data structure.

### 2.3.4 Fuzzy *c*-means clustering

Fuzzy *c*-means clustering (FCM) (Dunn, 1973) is a soft clustering approach where each data point can belong to more than one cluster. Each data point is assigned a weight (membership grade) indicating the degree to which the data point belongs to the cluster. We denote a matrix $W = \{w_{ij}\} \in [0,1]$, where $w_{ij}$ referring to the degree to which data point $x_i$ belongs to cluster $C_j$, $i = 1, \dots, n$; $j = 1, \dots, k$. The goal of the FCM algorithm is to minimize the following objective function:

$$\mathcal{L}(\mathcal{C}, \mathcal{D}) = \sum_{i=1}^{n} \sum_{j=1}^{k} w_{ij}^m \, dist(x_i, s_j)^2 \tag{2}$$

Where:

$$w_{ij}^m = \frac{1}{\sum_{h=1}^{k} \left( \frac{dist(x_i, s_j)}{dist(x_i, s_h)} \right)^{\frac{2}{m-1}}} \tag{3}$$

$m$ is the fuzzy parameter indicating the degree of fuzziness, $m \in (1, +\infty)$. The centroid $s_j$ of cluster $C_j$ is computed by:

$$s_j = \frac{\sum_{i=1}^{n} w_{ij}^m x_i}{\sum_{i=1}^{n} w_{ij}^m} \tag{4}$$

**Algorithm**:
1. Select the number of clusters $k$.
2. Assign weights randomly for data points
3. *Repeat*
    a. For each cluster, compute its new centroid.
    b. For each data point, compute its new weight.
*Until* convergence (the weights' change between two iterations is no more than a given $\epsilon$).

**Complexity**: The time complexity of fuzzy $c$-means is similar to $k$-means, i.e., $\mathcal{O}(tkn)$, where $t$ is the number of iterations (P. Zhang & Shen, 2018).

### 2.3.5 EM clustering

EM is a soft clustering approach to finding the maximum likelihood estimates of parameters in probabilistic models (Dempster et al., 1977). There are two important steps: *expectation* ($E$) and *maximization* ($M$). EM uses the finite Gaussian mixtures model to estimate parameters until convergence. Each cluster corresponds to one of $k$ probability distributions in the mixture. Similar to fuzzy $c$-means clustering, each data point is assigned a membership probability for each cluster (Jin & Han, 2010). The EM algorithm is summarized as follows:

**Algorithm**:

1. Initialize with two randomly placed Gaussians (mean and standard deviation).

2. Refine the parameters iteratively with two alternating steps until convergence:

   a. $E$ step: re-estimate the membership possibility for each instance based on the current model

   b. $M$ step: re-estimate the model parameters based on the new membership possibilities.

3. Assign data points to the cluster with their membership probability.

In the $E$ step, the membership possibility for each instance $x_i$ w.r.t. cluster $C_j$ (Tan et al., 2016) is computed by:

$$P(C_j|x_i) = \frac{P(x_i|C_j)P(C_j)}{\sum_j^k P(x_i|C_j)P(C_j)} \tag{5}$$

where $P(x_i|C_j)$ is the probability density function of the Gaussian distribution.

In the $M$ step, the probability of data points coming from the cluster $C_j$ (cluster density) is determined by:

$$P(C_j) = \frac{1}{N}\sum_{i=1}^{n} P(C_j|x_i) \tag{6}$$

Update centroid $s_j$ for each cluster $C_j$:

$$s_j = \frac{\sum_{i=1}^{n} x_i\, P(C_j|x_i)}{\sum_{i=1}^{n} P(C_j|x_i)} \tag{7}$$

and cluster covariances (distance computations) are determined as:

$$\Sigma_j = \frac{\sum_{i=1}^{n}(x_i - s_j)(x_i - s_j)'\, P(C_j|x_i)}{\sum_{i=1}^{n} P(C_j|x_i)} \tag{8}$$

**Complexity**: In the EM algorithm, for each iteration, distance computations require $\mathcal{O}(nk)$. Hence, the distance computation takes $\mathcal{O}(d^3)$ where $d$ is the dimension. Therefore, in general, the time complexity of the EM algorithm for each iteration is $\mathcal{O}(nkd^3)$.

### 2.3.6 SOM clustering

Self-organizing maps (SOM) (Kohonen, 1982) is a center-based clustering scheme. The idea of SOM clustering is for each data point $x_i \in \mathcal{D}, i = 1 \ldots n$ we aim to represent $x_i$ with $d$-dimensions through an output space with two dimensions. SOM clustering creates a two dimensions output space where each node in the output space is associated with a "weight" vector. The weight vector contains the position of the node in the input space (dataset), and the weight vectors will move toward the input data when the best matching unit (BMU) is found based on the Euclidean distance from the observed data point to all weight vectors. The summary of the SOM clustering algorithm is described as follows (Bação et al., 2005):

**Algorithm**: Apart from symbols described in Table 4, let $W$ be a grid of $m$ nodes, $w_{uv}$ ($u$ and $v$ are their coordinates of the grid); $t$ be the iteration; $\alpha(t) \in [0,1]$ be the learning rate factor; $r(t)$ be the radius of the neighborhood function $h(w_{uv}, w_{min}, t)$; $\alpha(t)$ and $r(t)$ decrease monotonically over time. The Gaussian is a commonly used neighborhood function (Natita et al., 2016):

$$h(w_{min}, w_{uv}, t) = \alpha(t)e^{\left(-\frac{dist(w_{min}, w_{uv})^2}{2r(t)^2}\right)} \tag{9}$$

1. Initialize the weight vectors of nodes in the output space randomly.
2. *Repeat*
   a. For each data point $x_i$:
      i. For all $w_{uv} \in W$ calculate $d_u v = \text{dist}(x_i - w_u v)$
      ii. Select the node that minimizes $d_{uv}$ as $w_{min}$
      iii. Update node $w_{uv} \in W$: $w_{uv} = w_{uv} + \alpha h(w_{min}, w_{uv}, t)dist(x_i, w_{uv})$
   b. Increase $t$
   *Until* the number of iterations $t$ is reached.

**Complexity**: The computation complexity of the SOM clustering is $\mathcal{O}(nmdt)$, where $n$ is the number of data points in the dataset $\mathcal{D}$, $m$ is the number of nodes of the output space, d is the dimension of each data point, and $t$ is the number of iterations (Melka & Mariage, 2017).

### 2.3.7 Spectral clustering

In spectral clustering (Shi & Malik, 2000; Ng et al., 2001), the spectrum (eigenvalues) of the similarity matrix of the data is used to reduce dimensionality before clustering in fewer dimensions. A similarity matrix is provided as the input and represents quantitative measurements of the similarity between two points. We summarize a well-known version of the spectral clustering proposed by Ng et al. (2001) as follows:

**Algorithm**: Given a dataset $\mathcal{D} = \{x_1, x_2, ..., x_n\}$ and $k$ is the number of clusters.

1. Create the similarity matrix $A \in \mathbb{R}^{n \times n}$ where $A_{ij} = e^{\left(-\frac{dist(x_i, x_j)^2}{2\sigma^2}\right)}$ if $i \neq j$, else $A_{ij} = 0$. The scaling parameter $\sigma^2$ is used to decide how quickly $A_{ij}$ falls off with the distance between $x_i$ and $x_j$.

2. Create the diagonal matrix $G$ with $G_{ii}$ is the sum of column $i$ of matrix $A$. Define the Laplacian $L = G^{-1/2} A G^{-1/2}$.

3. Find $k$ largest eigenvectors of $L$, denoted as $u_1, u_2, ..., u_n$. They should be orthogonal in the case of repeated eigenvalues. Then, we create the matrix $U = [u_1 u_2 ... u_k] \in \mathbb{R}^{n \times k}$ by columnarizing the eigenvectors.

4. Renormalize each row of $U$ to have unit length to create matrix $V$, where $V_{ij} = \frac{u_{ij}}{\sqrt{\sum_j u_{ij}^2}}$.

5. Consider each row of matrix $V$ as a point in $\mathbb{R}^k$ and use $k$-means (or any other clustering methods) to cluster them into $k$ clusters.

6. If and only if row $i$ of the matrix $V$ is assigned to cluster $j$, we assign the original point $x_i$ to cluster $j$.

**Complexity**: In comparison with other clustering methods, spectral clustering has a high computational complexity of $\mathcal{O}(n^3)$ in general. Therefore, spectral clustering is not suitable for large datasets (D. Yan et al., 2009).

## 2.3.8 DBSCAN clustering

DBSCAN is a density-based clustering that discovers the clusters and the noise in a spatial database (Ester et al., 1996). There are two parameters: *Eps* (the maximum radius of the neighborhood) and *MinPts* (the minimum number of points in an *Eps*-neighborhood of that point). In DBSCAN clustering, the points are categorized as *core points* (points inside of the cluster), *border points* (points on the border of the cluster) and *outliers*.

- A point $p$ is a core point if at least $minPts$ points are within distance *Eps* of $p$ (including $p$), i.e., $|N_{Esp}(p) = \{q \mid dist(p, q) \leq Eps\}| \geq MinPts$.
- A point $q$ is the border point if it has fewer than *MinPts* points within *Eps* radius, but it is in the neighborhood of a core point.
- *Outliers* are any points that cannot be reached from any other point.
- A point $p$ is directly density-reachable from point $q$ w.r.t. *Eps*, *MinPts* if $p \in N_{Eps}(q)$ and $q$ is a core point.

- A point $p$ is density-reachable from a point $q$ w.r.t. *Eps*, *MinPts* if there is a chain of points $p_1, ..., p_n$, where $p_1 = q, p_n = p$, such that $p_{i+1}$ is directly density-reachable from $p_i$.
- A point $p$ is density-connected to a point $q$ w.r.t. *Eps*, *MinPts* if there is a point $o$ such that both $p$ and $q$ are density-reachable from $o$ w.r.t. *Eps*, *MinPts*.

A cluster is a maximal set of density-connected points. Based on the above notations, the DBSCAN clustering can be summarized as follows:

**Algorithm**:

1. Start at a random point $p$.

- 2. Find all points density-reachable from $p$ w.r.t. *Eps*, *MinPts*.

3. If $p$ is a core point, form a cluster starting with $p$ and expand the cluster through neighbors of $p$.

4. If $p$ is a border point, and no point is density-reachable from $p$, DBSCAN visits the next point of the dataset.

5. Repeat the process until all points are processed.

**Complexity**: In general, $\mathcal{O}(n \times \text{time to find points in the Eps} - \text{neighborhood})$ is the runtime complexity. In the worst case, it is $\mathcal{O}(n^2)$. The average time complexity can be reduced to $\mathcal{O}(n\log n)$ by using an efficient data structure (e.g., kd-trees).

# 3 Fair clustering models (for EDS)

In this section, we find the answer to the research question $RQ_2$ by summarizing the popular fairness notions for clustering and well-known fair clustering models and investigating the applicability of those models in EDS.

## 3.1 Fairness notions

In general, the fairness notions depend on the application and specific context. There are 20 fairness notions used for fair clustering summarized in a survey of fairness in clustering (Chhabra et al., 2021). In their review, the fairness notions are categorized into four types: *group-level*, *individual-level*, *algorithm agnostic*, and *algorithm specific*. Based on the popularity of the fairness notions[9] (Chhabra et al., 2021), we summarize the following four notions: *balance*, *bounded representation*, *social fairness*, and *individual fairness*.

---

[9] We only take into account the fairness notions introduced in the published papers. Because the fairness notions may be turned into measures, therefore, in this review we use the term "fairness notion" and "fairness measure" interchangeably.

In the next sections, we denote $\mathcal{G}$ as the protected attribute with two values $\{F, M\}$, e.g., *gender = {female, male}*. In which, $F$ is the discriminated group (or protected group), e.g., "*female*", and $M$ is the non-discriminated group (or non-protected group), e.g., "*male*". Besides, we continue using the symbols listed in Table 4.

### 3.1.1 Balance

*Balance* is the most popular group-level fairness notion used in studies in fair clustering, which was introduced by Chierichetti et al. (2017). Given a clustering $\mathcal{C} = \{C_1, C_2, ..., C_k\}$ with $k$ clusters, the *balance* of a cluster $C_i$ is the minimum ratio between the number of objects in the protected group and the non-protected group. The *balance* of a clustering is the minimum *balance* score of all clusters. If each cluster has a balance of at least $\theta$ as defined by the balance requirement $\theta$, then clustering is fair. The *balance* measure can be applied in any fair clustering model.

$$balance(C_i) = min\left(\frac{\#F \in C_i}{\#M \in C_i}, \frac{\#M \in C_i}{\#F \in C_i}\right) \tag{10}$$

$$balance(\mathcal{C}) = min_{i=1}^{k}(C_i) \tag{11}$$

### 3.1.2 Bounded representation

*Bounded representation* is a generalization of disparate impact and was introduced by Ahmadian et al. (2019). This group-level notion aims to reduce imbalances in cluster representations of protected attributes (for example, gender). Let $\alpha$ be the over-representation parameter; a cluster $C_i$ is fair if the fractional representation of each group (protected, non-protected group) in the cluster is at most $\alpha$.

$$\frac{|\{x \in C_i | \mathcal{G} = g\}|}{|C_i|} \leq \alpha, \text{where} g \in F, M \tag{12}$$

Then, a clustering $\mathcal{C}$ is fair if all clusters satisfy the representation constraint. Similar to the *balance* notion, all fair clustering models can apply the bounded representation measure. Besides, *bounded representation* is generalized with two parameters $\alpha$ and $\beta$ in the study of Bera et al. (2019), where for each value of the protected attribute, the fractional representation of it in the cluster must be between $\beta$ and $\alpha$.

### 3.1.3 Social fairness

*Social fairness* was first introduced by Ghadiri et al. (2021), which aims to provide equitable costs for different clusters. In the *k*-means algorithm, the target is to minimize the objective function (recall Eq. 1).

$$\mathcal{L}(\mathcal{C}, \mathcal{D}) = \sum_{i=1}^{k} \sum_{x \in C_i} dist(x, s_i)^2$$

We denote $\mathcal{D}_F$ and $\mathcal{D}_M$ are two subsets of dataset $\mathcal{D}$, which contain values $F$ and $M$ of the protected attribute, respectively; $\mathcal{D} = \mathcal{D}_F \cup \mathcal{D}_M$. Then, the fair *k*-means objective is the larger average cost (social fairness cost):

$$\Phi(\mathcal{C}, \mathcal{D}) = max \left\{ \frac{\mathcal{L}(\mathcal{C}, \mathcal{D}_F)}{\mid \mathcal{D}_F \mid}, \frac{\mathcal{L}(\mathcal{C}, \mathcal{D}_M)}{\mid \mathcal{D}_M \mid} \right\} \tag{13}$$

The goal of the fair *k*-means is to minimize the social fairness cost $\Phi(\mathcal{C}, \mathcal{D})$. A disadvantage of this measure is that it can only be applied to center-based clustering because it is modeled on the objective function of a particular algorithm (Chhabra et al., 2021).

### 3.1.4 Individual fairness

Chakrabarti et al. (2022) introduced two individual-level fairness notions for *k*-center clustering, namely *Per-Point Fairness* and *Aggregate Fairness*. Given a dataset $\mathcal{D}$, the goal is to choose a set $S \subseteq \mathcal{D}$ of at most *k* centers and then find an assignment $\lambda: \mathcal{D} \mapsto S$. We denote $\alpha \geq 1$ is a parameter. For all data points $x \in \mathcal{D}$, $\mathcal{I}_x \subseteq \mathcal{D}$ is denoted as the group of data points that are similar to $x$. The fairness notions are described as follows:

**Per-Point Fairness**: For all data points $x \in \mathcal{D}$, *Per-Point Fairness* is satisfied if the distance from $x$ to its center is at most $\alpha$ time the distance of the closest point in that cluster to the center $\lambda(x)$.

$$dist(x, \lambda(x)) \leq \alpha \, min_{y \in \mathcal{I}_x}\{dist(y, \lambda(y))\} \tag{14}$$

**Aggregate Fairness**: For all data points $x \in \mathcal{D}$, *Aggregate Fairness* is satisfied if the distance from $x$ to its center is at most $\alpha$ time the *average* distance of the points in that cluster to the center $\lambda(x)$.

$$dist(x, \lambda(x)) \leq \alpha \frac{\sum_{y \in \mathcal{I}_x} dist(y, \lambda(y))}{|\mathcal{I}_x|} \tag{15}$$

## *3.2 Fair clustering models*

In this section, we overview the fair clustering models based on their popularity of the corresponding traditional clustering models. We hypothesize that if a clustering model is widely used in the education setting, its corresponding fair model will also be more applicable. Therefore, we focus on center-based clustering (such as $k$-means, $k$-center, $k$-medoids), hierarchical clustering, and spectral clustering.

### 3.2.1 Fair center-based clustering

**Fair $k$-center clustering**. The first work on group-level fair clustering was introduced by Chierichetti et al. (2017) to ensure an equal representation of each protected attribute in every cluster. They defined a new fairness notion, *balance*, described in Section 3.1.1. A two-phase approach was proposed: 1) Fairlet decomposition - clustering all instances into *fairlets* which are small clusters guarantying fairness constraint; 2) Applying vanilla clustering methods ($k$-center, $k$-median) on those *fairlets* to obtain the final resulting fair clusters. Fig. 3 illustrates the fair clustering method using the *fairlets* concept.

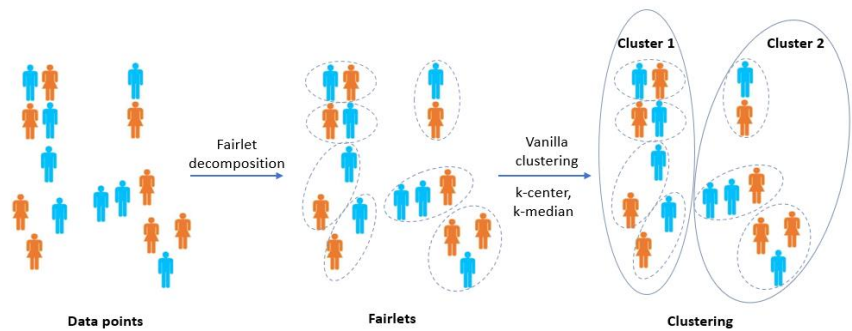

**Fig. 3** Fair clustering through fairlets.

Besides, Ahmadian et al. (2019) introduced a new fairness measure (*bounded representation*) by providing an upper bound constraint for fairness in resulting clusters, applied for $k$-center clustering. Jones et al. (2020) proposed an algorithm with a linear time complexity and obtained a 3-approximation for the fair $k$-center summaries problem. Recently, Chakrabarti et al. (2022) presented two individual fairness notions that guarantee each data point has a similar *quality of service*. They proposed approximation algorithms for the $k$-center objective.

**Fair $k$-means clustering**. Schmidt et al. (2019) proposed a fair $k$-means clustering by using the *coresets* concept. Essentially, a *coreset* is a summary of a point set that approximates any possible solution's cost function well. They extended the approach of Chierichetti et al. (2017) to compute the fairlet for *k*-means algorithm. Their experiments show that the new method can be applied to big data. Besides, Ghadiri et al. (2021) introduced the social fairness notion, which focuses on minimizing the clustering cost across groups of the protected attribute. They proposed a fair clustering version of the well-known Lloyd *k*-means and reported results through clustering cost. Abraham et al. (2020) introduced a fair *k*-mean clustering model, namely *FairKM*, which combines the optimization of the classical clustering objective and a novel fairness loss term. Their model aims to achieve a trade-off between clustering quality (on the non-protected attributes) and cluster fairness (on the multiple protected attributes).

**Fair *k*-medoids clustering**. *k*-medoids clustering and fairlet decomposition were applied in the study of Le Quy et al. (2021). They introduced the grouping problem in the educational setting, where each cluster has a limitation in terms of cardinality, and the clustering must be fair w.r.t. protected attribute. They applied the fairlet decomposition (Chierichetti et al., 2017) to obtain the fairness w.r.t. protected attribute, and *k*-medoids clustering was used to satisfy the cardinality constraint. To the best of our knowledge, this is the first work taking into account fairness and clustering in the educational environment.

**Fair fuzzy *c*-means clustering**. A fair fuzzy *c*-means clustering was proposed by Xia et al. (2021). Their goals are to minimize the objective function $\mathcal{L}(\mathcal{C}, \mathcal{D})$ (Eq. 2), and maximize the ratio of each group (e.g., *female, male*), i.e., as close as possible to the ratio in the original dataset. They defined a new *fair loss function* to measure the fair loss in clusters and optimized the assignment phase according to the objective function to obtain a fair clustering.

### 3.2.2 Fair hierarchical clustering

Ahmadian et al. (2020) defined a fair hierarchical clustering for any fairness constraint, in which "a hierarchical clustering is fair if all of its clusters (besides the leaves) are fair". They extended the fairlet decomposition (Chierichetti et al., 2017) for upper-bounded representation fairness and proved the results for the revenue, value, and cost objectives. In the educational domain, Le Quy et al. (2021) proposed a fair-capacitated clustering and used hierarchical clustering to obtain the clusters with a given maximum capacity, while the fairness constraint was satisfied by using the fairlet decomposition approach.

### 3.2.3 Fair spectral clustering

Kleindessner et al. (2019) incorporated the balance fairness notion in normalized and unnormalized constrained spectral clustering. They applied *k*-means cluster-ing on a fair subspace generated by projecting the graph Laplacian. As mentioned in Section 2.3.7, spectral clustering is used widely in educational data analysis. Therefore, the fair version of this model is a potential candidate for EDS.

# 4 (Fair) clustering evaluation in EDS

In this section, we investigate the evaluation aspects of clustering models toward fairness, including evaluation measures, datasets and parameter selection to tackle the research question $RQ_3$.

## 4.1 Evaluation measures

### 4.1.1 Clustering quality measures

Table 5 outlines the measures used in related work to evaluate the cluster validity. We summarize the most prevalent measures as follows:

**Silhouette coefficient**. This is a metric that measures cluster separation and compactness at the same time for both individual data points, as well as clusters and clustering. The silhouette score of a data point $x_i$ in cluster $C_I$ is computed by:

$$score_i = \frac{\beta_i - \alpha_i}{max(\alpha_i, \beta_i)} \tag{16}$$

and $score_i = 0$ if $\mid C_I \mid = 1$.

In Eq. 16, $\alpha_i$ is the average distance of the data point $x_i$ to other points in its cluster $C_I$, i.e., $\alpha_i = \frac{\sum_{j \neq i, x_j \in C_I} dist(x_i, x_j)}{|C_I| - 1}$, and $\beta_i$ is the minimum of the average distance of the point $x_i$ to points in other clusters, $\beta_i = min_{J \neq I} \frac{1}{|C_J|} \sum_{x_j \in C_J} dist(x_i, x_j)$. Finally, the silhouette coefficient of the entire dataset is the maximum value of the mean $score_i$ over all data points (Kaufman & Rousseeuw, 1990).

**Sum of Squared Error (SSE) or cluster cohesion**. SSE measures how closely related objects are in a cluster, which is computed by objective function:

$$SSE = \sum_{i=1}^{k} \sum_{x \in C_i} dist(x, s_i)^2 \tag{17}$$

where $k$ is the number of clusters, and $s_i$ is the centroid of the cluster $C_i$.

**Davies–Bouldin index (DBI)** is an internal evaluation scheme introduced by Davies and Bouldin (1979). DBI is calculated by the following formula:

$$DBI = \frac{1}{k} \sum_{i=1}^{k} max_{j \neq i} \left( \frac{\sigma_i + \sigma_j}{dist(s_i, s_j)} \right) \tag{18}$$

where $k$ is the number of clusters, $s_i$ is the centroid of cluster $i$-th, $\sigma_i$ is the average distance of all cluster members in cluster $i$ to its centroid $s_i$.

**Dunn index (DI)** was introduced by Dunn (1974), which identifies clusters that are well-separated and dense. DI is computed by dividing the minimal inter-cluster distance by the maximal intra-cluster distance.

$$DI = \frac{min_{1 \leq i < j \leq k} dist(i, j)}{max_{1 \leq p \leq k} dist'(p)} \tag{19}$$

where $dist(i, j)$ is the distance between clusters $i$-th and $j$-th (e.g., the distance between two centroids), and $dist'(p)$ is the intra-cluster distance of cluster $p$ (e.g., the maximal distance between any pair of data points in cluster $p$).

**Table 5** Clustering quality measures

| Measures | Used in |
|---|---|
| Silhouette coefficient | Battaglia et al. (2017); Bharara et al. (2018); Fang et al. (2018); Khayi and Rus (2019); Howlin and Dziuban (2019); Abraham et al. (2020); Maylawati et al. (2020); Moubayed et al. (2020); Šarić-Grgić et al. (2020); Ahmed et al. (2021); X. Li et al. (2021); Palani et al. (2021); Xia et al. (2021); T. Zhang et al. (2022); Talebinamvar and Zarrabi (2022); Hassan et al. (2022). |
| Sum of Squared Error (SSE) | Kurniawan et al. (2018); Schmidt et al. (2019); Rijati et al. (2018); Abraham et al. (2020); W. Chang et al. (2020); Le Quy et al. (2021); Xia et al. (2021); Yin (2021); Sobral and de Oliveira (2021). |
| Davies–Bouldin index (DBI) | W. Chang et al. (2020); X. Li et al. (2021); Ahmed et al. (2021); Palani et al. (2021); Delgado et al. (2021). |
| Calinski–Harabasz index (CHI) | C. Wang & Zha (2018); W. Chang et al. (2020); X. Li et al. (2021); Palani et al. (2021). |
| ANOVA test | Shen and Chi (2017); Aghababyan et al. (2018). |
| Dunn index (DI) | Fang et al. (2018); Ahmed et al. (2021). |
| Normalized mutual information (NMI) | Vo and Nguyen (2018); Q. Tang et al. (2021). |

| Measures | Used in |
|---|---|
| Adjusted Rand Index (ARI) | Mishler and Nugent (2018). |
| Chi-Square | Wu et al. (2021). |
| Clustering accuracy (ACC) | Q. Tang et al. (2021). |
| Fuzzy silhouette index | Howlin and Dziuban (2019). |
| Heterogeneity score | Wu et al. (2021). |
| Homogeneity score | Wu et al. (2021). |
| Log-likelihood | Wu et al. (2021). |
| Modified partition coefficient | Howlin and Dziuban (2019). |
| Partition coefficient | Howlin and Dziuban (2019). |
| Partition entropy | Howlin and Dziuban (2019). |
| Simpson index | C. Wang and Zha (2018). |
| t-tests | Mojarad et al. (2018). |
| Xie and Beni index | Howlin and Dziuban (2019). |

**4.1.2 Fairness measures for fair clustering**

Balance is the most prevalent fairness measure used in the fair clustering models (Chierichetti et al., 2017; Kleindessner et al., 2019; Schmidt et al., 2019; Le Quy et al., 2021). Xia et al. (2021) reported the clustering results in terms of fairness by the following measures: *balance* (Bera et al., 2019)*, Euclidean distance of distribution vectors* and *Wasserstein distance of distribution vectors*. Abraham et al. (2020) used the following fairness measure to evaluate their model: *Average Euclidean, Average Wasserstein, Max Euclidean, Max Wasserstein*. Besides, Chakrabarti et al. (2022) reported their experimental results on two fairness measures: *Per-Point Fairness* and *Aggregate Fairness*. In terms of clustering cost, Ghadiri et al. (2021) presented their result on *social fairness*, which is the clustering cost of each group of the protected attribute. In addition, the *bounded representation* measure was used to report the resulting clustering for fair hierarchical clustering (Ahmadian et al., 2020).

## *4.2 Datasets*

In the related work, most datasets are non-public. The data are collected by the paper's authors or shared within their institute. Therefore, accessing educational datasets is a significant challenge. This section summarizes the educational datasets used in the references w.r.t. (fair) clustering. Due to the capacity of the book chapter, we only list the datasets used in the top journals and conferences, i.e., ranking Q1, A*/A. In Table 6, we outline 27 non-public educational datasets used in the related work. Besides, seven public educational datasets are summarized in Table 7. The datasets are sorted in ascending order of the paper's publication year. The majority of the datasets are tabular and collected in MOOCs and e-learning platforms. Most of the datasets are small, and only seven datasets contain more than 10,000 instances.

**Table 6** Non-public datasets used to evaluate (fair) clustering models.

| Dataset & Description | Used in | Protected attributes | Data source | Data type |
|---|---|---|---|---|
| 427,382 logs in the time period of 16 weeks and 546,966 cleaned logs in 18 courses on Moodle online graduate program in the United States | Hung et al. (2017) | Ethnicity, gender | Online learning platform | Tabular |
| Course "Social Aspects of Information Technology" in the iMooX platform with 459 matriculated under-graduates and 379 external students | Khalil and Ebner (2017) | Gender | MOOC | Tabular |
| 811 records covering the admission years of 2007–2009 at Mae Fah Luang University | Iam-On et al. (2017a) | Gender | Traditional classroom | Tabular |
| 1,199 university students and 35 teachers (277 students and 19 teachers are selected for experiments) | Urbina Nájera et al. (2017) | Gender, marital status | Traditional classroom | Tabular |
| MITx introductory programming course with 12,973 correct and 4,293 incorrect attempts | Gulwani et al. (2018) | - | MOOC | Text |
| 1,093 students in two e-tutorials: SOWISO and MyStatLab (MSL) in 2016/2017 | Tempelaar et al. (2018) | Gender | Online learning platform | Tabular |
| Synthetic dataset of 600 students | Kausar et al. (2018) | - | Synthetic | Tabular |
| 438 and 617 Chinese public universities in 1998 and 2011, respectively | C. Wang et al. (2018) | - | Traditional classroom | Tabular |
| 64 students in projects in a graduate software engineering course | Akbar et al. (2018) | - | MOOC | Tabular |
| 10 million meals w e r e consumed by about 82,871 students at the canteen of the University of Pisa | Natilli et al. (2018) | Gender | Canteen of university | Tabular |
| Real student dataset of various academic disciplines of higher educational institutions in Kerala, India | Francis et al. (2019) | Gender | Traditional classroom | Tabular |

| Dataset & Description | Used in | Protected attributes | Data source | Data type |
|---|---|---|---|---|
| Three groups of learners (size 30, 50, 80) in Hubei province of China (National Training Plan 2015) | H. Liu et al. (2019) | - | Traditional classroom | Tabular |
| 486 students attending undergraduate science courses at the University of Western Ontario in Canada | Moubayed et al. (2020) | - | Online learning platform | Tabular |
| 102 middle school students in the Northeastern United States | S. Li et al. (2020) | Race | Traditional classroom | Tabular |
| 1,062 students from Balqa Applied University (BAU) in Jordan from 2006 to 2018 | Almasri et al. (2020) | Gender | Traditional classroom | Tabular |
| 2,000 students from 4 universities in China | W. Chang et al. (2020) | Gender | Traditional classroom | Tabular |
| Two online Python programmingcourses: intermediate (42,131 students) and beginners (7,164 student) in Australia | McBroom et al. (2020) | - | Online learning platform | Tabular |
| 9,024 undergraduates at a university in Beijing during the spring of 2019 | X. Li et al. (2021) | - | Traditional classroom | Tabular |
| 180 students | Yin (2021) | Gender | Traditional classroom | Tabular |
| 1,709,189 records of online students enrolled from 2015 to 2019 at Universidad Internacional de La Rioja (UNIR) | Delgado et al. (2021) | - | Online learning platform | Tabular |
| 8,201 feedback responses for 168 distinct courses from the University Politehnica of Bucharest in 2019–2020 | Masala et al. (2021) | - | Survey | Text |
| 121 students in programming didactic course at Stockholm University in 2020 | Wu et al. (2021) | - | Traditional classroom | Text |
| Programming course in online learning environment, training: 11 problems (2,358 solutions), test: 11 new problems (2,598 solutions) | Effenberger et al. (2021) | - | Online learning platform | Text |
| Two datasets: 29 students from the international relations course and 50 students from the computer science and computer engineering courses | Sobral et al. (2021) | Gender | Traditional classroom | Tabular |
| 251 students' click-stream log data in an introductory physics course (Fall 2020 semester) | T. Zhang et al. (2022) | - | Online learning platform | Text |
| 180 male and 458 female Iranian B.A. students involving 20 classes at Mazandaran University of | Tale-binamvar et al. (2022) | Gender | Online learning platform | Tabular |

| Dataset & Description | Used in | Protected attributes | Data source | Data type |
|---|---|---|---|---|
| Science and Technology | | | | |
| 400 first-year students in a Medium sized Metropolitan University in Dublin with two programming courses | Mai et al. (2022) | - | Online learning platform | Tabular |

**Table 7** Public datasets used to evaluate (fair) clustering models.

| Dataset & Description | Used in | Protected attributes | Data source | Data type |
|---|---|---|---|---|
| Dataset of shell commands [10] with18 cybersecurity training sessions, 8,834 commands collected from 113 trainees | Švábenský` et al. (2022) | - | Online learning platform | Text |
| 480 students collected by Kalboard 360 LMS (xAPI- Edu-Data[11]) | Bharara et al. (2018) | Gender | Online learning platform | Tabular |
| OULAD dataset [12] with 32,593 students enrolled in 7 courses at Open university in England | Casalino et al. (2019); Le Quy et al. (2021); Palani et al. (2021) | Gender | Online learning platform | Tabular |
| Student performance dataset[13] of two Portuguese secondary schools | Jones et al. (2020); Le Quy et al. (2021) | Gender | Traditional classroom | Tabular |
| PISA test scores dataset [14] with information of 5,233 American students in 2009 from the Program for International Student Assessment (PISA) | Le Quy et al. (2021) | Gender | Traditional classroom | Tabular |
| MOOC dataset[15] of 416,921 students in the MOOCs enrolled in the 16 edX courses offered by Harvard and MIT during 2012-2013 | Le Quy et al. (2021) | Gender | MOOC | Tabular |

---

[10] https://gitlab.ics.muni.cz/muni-kypo-trainings/datasets/commands

[11] https://www.kaggle.com/datasets/aljarah/xAPI-Edu-Data

[12] https://analyse.kmi.open.ac.uk/open_dataset

[13] https://archive.ics.uci.edu/ml/datasets/student+performance

[14] https://www.kaggle.com/econdata/pisa-test-scores

[15] https://github.com/kanika-narang/MOOC_Data_Analysis

| Dataset & Description | Used in | Protected attributes | Data source | Data type |
|---|---|---|---|---|
| StudentLife dataset[16] | Hassan et al. (2022) | - | Online learning platform | Tabular |

## *4.3 Parameter selection*

### 4.3.1 Center-based clustering

*k*-**means clustering**. To choose the best value for the number of clusters $k$, researchers run the clustering with several values for $k$ and select the one that does not result in a significant improvement in the compactness of the cluster at any further increase in $k$ (Natilli et al., 2018). In the study of Yang (2021), $k$ was determined based on the experimental analysis of the real application of clustering results for students' development, optimizing education and teaching. In another approach, researchers determined the optimal $k$ based on 10-fold cross-validation implemented on the training set (Adjei et al., 2017). Besides, the silhouette coefficient is also used as an essential criterion to select the best number of clusters (Bharara et al., 2018; Moubayed et al., 2020; X. Li et al., 2021; Hassan et al., 2022). The pre-implemented package, namely *NbClust*[17] is also used to find the optimal $k$ (Khalil & Ebner, 2017; Oladipupo & Olugbara, 2019). In addition, T. Zhang et al. (2017) proposed a method to select the initial center points by partitioning the dataset into $k$ domains and choosing the points with the greatest density in each domain as the initial cluster centers. Aghababyan et al. (2018), Nazaretsky et al. (2019) and Ahmed et al. (2021) applied within the SSE between data points in clusters to select the number of clusters. In the fair $k$-means version, Schmidt et al. (2019) experimented their proposed method with the number of clusters in a range [20, 100], and Ghadiri et al. (2021) used the range [4,16] for $k$.

*k*-**medoids clustering**. Regarding $k$-medoids clustering, the Euclidean distance is usually used as the dissimilarity measure (Furr, 2019), and the number of medoids $k$ is determined by using the average silhouette coefficient (Kausar et al., 2018; Furr, 2019). In the study of Le Quy et al. (2021) with the fair-capacitated $k$-medoids model, the number of clusters is experimented with in a range of [3, 20].

*k*-**cencer fair clustering**. Chakrabarti et al. (2022) performed experiments with the number of clusters $k$ in {2, 4, 8, 16, 32, 64, 128}. A set of $k$ in {3, 4, 6, 8, 10, 12, 14, 16, 18, 20} was selected for experiments in the work of Chierichetti et al. (2017). However, the optimal value of $k$ was not discussed in their studies.

---

[16] https://studentlife.cs.dartmouth.edu/dataset.html

[17] https://cran.r-project.org/web/packages/NbClust/index.html

### 4.3.2 Hierarchical clustering

In the related work, Euclidean distance is a similar common dissimilarity measure (Kylvaja et al., 2019; S. Li et al., 2020; Yotaman et al., 2020; T. Zhang et al., 2022). Yotaman et al. (2020) used Manhattan distance and several linkage algorithms: average link, complete linkage, single linkage, Ward's linkage, and median linkage in their experiments. However, Ward's linkage is the popular linkage applied for proximity matrix computation (Shen & Chi, 2017; Kylvaja et al., 2019; S. Li et al., 2020; T. Zhang et al., 2022). The number of clusters is chosen by using the local maximum average (T. Zhang et al., 2022) or based on the meaning of the resulting clusters (Kylvaja et al., 2019). In other approaches, the gap statistics (comparing total intra-cluster variation against expected values under distributions exhibiting no obvious clustering patterns) are applied to evaluate the resulting clusters and determine the number of clusters (S. Li et al., 2020). The knee position, which is described by the local change of slope of the SSE significantly, is applied to identify the number of clusters (Shen & Chi, 2017; Yotaman et al., 2020). Le Quy et al. (2021) varied the number of clusters between 3 and 20 to experiment with their fair-capacitated hierarchical clustering model.

### 4.3.3 Fuzzy *c*-means clustering

The parameters are set by a heuristic strategy, with a threshold $\epsilon = 10e\text{-}5$ (Q. Tang et al., 2021). Usually, the fuzzy parameter is set at $m = 2$ (Howlin & Dziuban, 2019; Amalia et al., 2019; Q. Tang et al., 2021). Besides, the number of clusters is set in a range, and the optimal $k$ is determined by using the gap statistics method (Palani et al., 2021) or based on the variation trend of the error of the predictive model (F. Liu, 2021). Howlin and Dziuban (2019) used validity indices implemented in *FClust*[18] R package to select the optimal $k$. In addition, Oladipupo and Olugbara (2019) applied *NbClust* package to find the number of clusters. In the fair fuzzy *c*-means clustering, the number of clusters $k$ was chosen in {4, 6, 8} (Xia et al., 2021).

### 4.3.4 EM clustering

T. Zhang et al. (2022) used the R package *mixsmsn*[19] to execute the EM clustering (with fitting finite mixtures of uni- and multivariate scale mixtures of skew-normal distributions) for the categorization of events. A probabilistic mixture model with Pois- son distribution within the EM algorithm is applied to cluster students' procrastination behavior (Park et al., 2018). They used the Bayesian approach and Gamma prior distributions for the rate parameters to fit the mixture model with two clusters ($k = 2$) on the student's dataset. In addition, the Log-likelihood measure and Schwarz's Bayesian criterion are used for EM clustering in the study of Ruipérez-Valiente et al. (2017), with the number of clusters set at $k$ = 3 based on cluster quality regarding cohesion and separation. With an assumption that the resulting clusters of students should not exceed 20 members, Urbina Nájera et al. (2017) applied the EM to divide students into $k = 14$ groups. Preetha (2021) set the number of clusters at $k = 4$ to group students based on their performance. Moreover, *NbClust* pre-implemented package is also used to determine the number of clusters (Oladipupo & Olugbara, 2019).

---

[18] https://cran.r-project.org/web/packages/fclust/index.html

[19] https://cran.r-project.org/web/packages/mixsmsn/index.html

### 4.3.5 SOM clustering

To determine the number of nodes of the output space, Delgado et al. (2021) trained different SOM models with a different number of nodes and selected the one with the lowest value of the topographic function (the average number of left-over neighborhood connections per output node). Besides, the number of clusters is chosen by using the connectivity indices *Conn_Index* (Tasdemir & Merényi, 2011) and DBI methods. The quality of clustering is evaluated by silhouette coefficient, DBI and DI measures (Alias, Ahmad, & Hasan, 2017). In addition, the number of nodes could be chosen as $m = \sqrt[3]{n}$ or equal to the number of clusters (Bara, Ahmad, Modu, & Ali, 2018). Ahmad et al. (2019) set the number of nodes as $m = 5\sqrt{n}$ and the neighborhood radius $r = \frac{max(m)}{4}$ after training 10 times.

### 4.3.6 Spectral clustering

In the related work, the cluster size is determined by $\sqrt{\frac{n}{2}}$ (Wu et al., 2021). Besides, the Elbow method is used to find the optimal number of clusters (Hooshyar et al., 2019) by comparing the distortion score with $k$ in a range of [2,4]. Kleindessner et al. (2019) performed the experiments with the number of clusters ranging from 2 to 8 and 1 to 15 for two datasets.

### 4.3.7 DBSCAN clustering

A heuristic method is used to find the optimal *MinPts* (X. Li et al., 2021) by using the graphs w.r.t. different *MinPts*, where the graph is the visualization of a function mapping each sample in the dataset to the distance from its *MinPts*-th nearest neighbor. Optimal *MinPts* is determined by selecting a minimum value whose *MinPts*-dist graph exhibits no significant difference from the rest. As a result, a maximal *E ps* can be obtained by finding the *MinPts*-dist value of the sample in the first "valley" in the graph of optimal *MinPts*. In the other work (Kausar et al., 2018), the $k$-nearest neighbor distances (the average distance of every data point to its $k$-nearest neighbors) in a matrix of data points are applied to determine the optimal value of *Eps* and *MinPts* parameters. In the study of Du et al. (2021), they computed the $k$-distance of each data point and then sorted these distances in ascending order before visualizing them in a scatter chart. *Eps* value is decided by the value of $k$-distance corresponding to the position that changes sharply. The parameter *MinPts* is chosen based on the minimum number of students with higher similarity in the class.

## 5 Beyond fairness requirements for clustering in EDS

In this section, we investigate other problems and needs apart from fairness for clustering in EDS to tackle the research question $RQ_4$.

**Cardinality constraints**. In collaborative learning, it is important to have comparable group sizes to distribute work fairly. However, many traditional and fair clustering models do not consider the cardinality of the resulting clusters. As a result, clustering models may produce groups of students of various sizes, which makes it difficult for teachers to assign work to groups. Therefore, clustering models need to ensure fairness and consider the resulting clusters' size to increase the usefulness and actionability of the clustering. The capacity constraint can be expressed as the minimum size, the maximum size of the clusters, or both (Mulvey & Beck, 1984; Bradley et al., 2000). Recently, the fair-capacitated clustering problem was introduced by Le Quy et al. (2021), which ensures fairness and balanced cardinalities of the resulting clusters. They proposed two solutions to achieve the capacitated clustering on the top of fairlet decomposition, namely hierarchical-based and $k$-medoid-based approaches.

**Explainability**. Explaining decisions is becoming increasingly important in education, especially in ML-based learning systems. ML algorithms, e.g., clustering models, are black boxes for decision-makers. However, because clustering algorithms rely on all data features, they produce cluster assignments that are difficult to explain. Hence, a fairly obvious requirement is that clustering models should provide explanations for the model's results, i.e., cluster assignments, in a way that humans can understand. Several studies are focusing on the explainability of well-known clustering models, such as $k$-means, $k$-medians (Moshkovitz et al., 2020), sby using a small decision tree. Bandyapadhyay et al. (2022) investigated the computational complexity of several variants of explainable clustering and introduced a new model of explainable clustering by a threshold tree after re-moving a subset of points. As mentioned above, fair clustering models aim to find the trade-off between the clustering quality and fairness in resulting clusters. Therefore, the goal of explainable models is to provide an explanation for these fair clustering models in terms of both cluster quality and fairness.

**Homogeneous and heterogeneous clustering**. In educational settings, both homogeneous and heterogeneous groups play an essential role. Typically, the objective of clustering models is to divide objects into homogeneous clusters, for example, students with the same ability level. Hence, homogeneous groups are the base for providing adequate support in a learning situation and allowing ability-specific learning. Besides, in collaborative learning, the heterogeneous groups w.r.t. characteristics and level of ability of students are more important to prevent critical group building (winners and looser, the smart students and the other) or to allow learning in teams with different specific abilities (Watson & Marshall, 1995; D.-Y. Wang et al., 2007; Pratiwi et al., 2017). In this way, collaborative learning can benefit from psychological factors in heterogeneous groups. Therefore, clustering algorithms need to consider the specific EDS task to determine whether the algorithm's output is homogeneous or heterogeneous clusters.

# 6 Conclusions and Outlook

Clustering techniques play a crucial role in analyzing students' performance, supporting learning, giving feedback and recommendation, and many EDS tasks. Studies on clustering algorithms and fairness in EDS can employ a variety of approaches and discussions.

First, in this chapter, we summarize the popular EDS tasks using clustering techniques and present the most prevalent clustering algorithms (traditional and fairness-aware models). We also discuss evaluation aspects w.r.t. clustering performance, datasets and fairness measures. In which, improving clustering performance is an essential requirement of the clustering models because the clustering quality is the objective of any clustering algorithm. Hence, by increasing clustering quality, the clustering models are also more likely to find meaningful clusters, which are useful for cluster analysis. In addition, fair clustering models also need to take into account the trade-off between clustering quality and the fairness of the resulting clusters.

Second, because EDS has grown rapidly due to the emergence of both novel data and innovative methods (McFarland et al., 2021), it is crucial to collect and develop a benchmark educational dataset for EDS. On the one hand, the vast majority of datasets are non-public and challenging to access due to their privacy issues, as demonstrated in our review. On the other hand, they are collected for different analytical purposes or problems. Therefore, the scope of use of these datasets is somewhat limited. Besides, generating the synthetic dataset for EDS is a potential solution to overcome the difficulties caused by the lack of benchmark educational datasets (Flanagan et al., 2022; Vie et al., 2022).

Third, with the rapid development of MOOC systems, researchers can collect and process large volumes of data in various formats, both structured and unstructured. Therefore, traditional clustering and corresponding fairness-aware models also need to be improved and upgraded to be able to handle such big educational data. For example, (fair) clustering methods should be scalable with big data (Fahad et al., 2014; Backurs et al., 2019) or be able to process high-dimensional data (Assent, 2012).

Forth, bias and discrimination are common problems in education. In this chapter, we study the well-known fairness measures applied in clustering models in various domains. However, because fairness notions differ across disciplines, it isn't easy to evaluate the efficiency of fairness-aware clustering algorithms. Therefore, selecting or defining the appropriate fairness notions for the educational domain is crucial and necessary. There are several studies on evaluation fairness for classification algorithms in education (Gardner et al., 2019; Le Quy et al., 2022a). However, choosing the appropriate fairness measures for fair clustering models in EDS is still a significant challenge for researchers.

Fifth, because clustering models are widely applied in many EDS tasks, and each problem has different constraints or requirements, constrained clustering models are also applied. For example, the recommendation systems take into account students' preferences or clustering models that support collaborative learning concern the groups' size. Moreover, the explainability and interpretability of (fair) clustering models are also interesting topics for future research.

To conclude, the construction and selection of clustering models (traditional or fair) in EDS is an important issue in improving the efficiency of educational data analysis and contributing to improving current educational systems. We overview the role of (fair) clustering models in EDS and investigate the use of existing clustering models in EDS. We believe that our survey can help both educational scientists and computer scientists have a broad picture of clustering models and fairness in education, then they can determine the appropriate clustering methods for their studies.

**Acknowledgments.** The work of the first author is supported by the Ministry of Science and Culture of Lower Saxony, Germany, within the Ph.D. program "LernMINT: Data-assisted teaching in the MINT subjects".

# 7 Appendix

**Table 8** Clustering methods used in EDS

| Clustering methods | #Papers | References |
|---|---|---|
| $k$-means | 73 | Adjei et al. (2017); Battaglia et al. (2017); Ding et al. (2017); Iam-On and Boongoen (2017a); Jia et al. (2017); Khalil and Ebner (2017); Hung et al. (2017); López et al. (2017); Salwana et al. (2017); Rihák and Pelánek (2017); Roy et al. (2017); Urbina Nájera et al. (2017); T. Zhang et al. (2017); Aghababyan et al. (2018); Akbar et al. (2018); Bharara et al. (2018); Esnashari et al. (2018); Fang et al. (2018); Kurniawan et al. (2018); Kausar et al. (2018); Gunawan et al. (2018); Mengoni et al. (2018); Mishler and Nugent (2018); Mojarad et al. (2018); Ninrutsirikun et al. (2018); Natilli et al. (2018); Nen-Fu et al. (2018); Purba et al. (2018); Tempelaar et al. (2018); Rijati et al. (2018); Vo and Nguyen (2018); C. Wang and Zha (2018); Dovgan et al. (2019); Francis and Babu (2019); L. Huang et al. (2019); Nazaretsky et al. (2019); Oladipupo and Olugbara (2019); Phanniphong et al. (2019); X. Wang et al. (2019); Waspada et al. (2019); Almasri et al. (2020); W. Chang et al. (2020); Chaves et al. (2020); Kosztyán et al. (2020); Maylawati et al. (2020); Moubayed et al. (2020); Pradana et al. (2020); Qoiriah et al. (2020); Shelly et al. (2020); P. Tang et al. (2020); Rijati et al. (2020); Šarić-Grgić et al. (2020); Ahmed et al. (2021); Chi (2021); Y.-W. Chu et al. (2021); Iatrellis et al. (2021); G. Li et al. (2021); X. Li et al. (2021); Masala et al. (2021); Preetha (2021); Putra et al. (2021); Sobral and de Oliveira (2021); Susanto et al. (2021); Rauthan et al. (2021); Q. Wang (2021); Yang (2021); Yin (2021); S. Zhang et al. (2021); Y. Guo et al. (2022); Hassan et al. (2022); Talebinamvar and Zarrabi (2022); Vasuki and Revathy (2022) |
| Hierarchical clustering | 31 | López et al. (2017); Salwana et al. (2017); Shen and Chi (2017); Rahmat (2017); Rihák and Pelánek (2017); Akbar et al. (2018); Fang et al. (2018); Mengoni et al. (2018); Mishler and Nugent (2018); Kausar et al. (2018); C. Wang and Žha (2018); Cheng and Shwe (2019); Dovgan et al. (2019); Kylvaja et al. (2019); Nazaretsky et al. (2019); Oladipupo and Olugbara (2019); S. Li et al. (2020); Pradana et al. (2020); Silva and Silla (2020); Singelmann et al. (2020); Popov et al. (2020); Yotaman et al. (2020); L.-H. Chang et al. (2021); Y. Chu and Yin (2021); J. B. Huang et al. (2021); Pamungkas et al. (2021); Preetha (2021); Mai et al. (2022); M. Wang and Lv (2022); T. Zhang et al. (2022) |
| Fuzzy $c$-means | 17 | Güvenç and Çetin (2018); Amalia et al. (2019); Howlin and Dziuban (2019); Casalino et al. (2019); Oladipupo and Olugbara (2019); Ramanathan et al. (2019); Varela et al. (2019); Supianto et al. (2020); |

| Clustering methods | #Papers | References |
|---|---|---|
| | | Thilagaraj and Sengottaiyan (2020); Yadav (2020); Y. Li et al. (2021); F. Liu (2021); Palani et al. (2021); Parvathavarthini et al. (2021); Q. Tang et al. (2021); Y. Li and Sun (2022); Premalatha and Sujatha (2022) |
| Distinctive clustering methods | 9 | Pratiwi et al. (2017); Gulwani et al. (2018); Fasanya and Fathizadeh (2019); Nguyen and Vo (2019); Waluyo et al. (2019); McBroom et al. (2020); Effenberger and Pelánek (2021); Z. Wang and Wang (2022); Y. Wang and Wang (2022) |
| EM | 6 | Ruipérez-Valiente et al. (2017); Urbina Nájera et al. (2017); Mengoni et al. (2018); Park et al. (2018); Oladipupo and Olugbara (2019); Preetha (2021); T. Zhang et al. (2022) |
| SOM clustering | 5 | Alias et al. (2017); Salwana et al. (2017); Bara et al. (2018); Ahmad et al. (2019); Purbasari et al. (2020); Delgado et al. (2021) |
| Spectral clustering | 5 | Mengoni et al. (2018); Dovgan et al. (2019); Gao et al. (2019); Hooshyar et al. (2019); Wu et al. (2021) |
| DBSCAN | 4 | Kausar et al. (2018); Oladipupo and Olugbara (2019); Du et al. (2021); X. Li et al. (2021) |
| $k$-medoids | 3 | Kausar et al. (2018); Furr (2019); Vasuki and Revathy (2022) |
| Affinity propagation | 1 | H. Liu et al. (2019) |
| Author-Topic Model | 1 | Rakhmawati et al. (2021) |
| Bio-inspired clustering | 1 | M.-H. Chang et al. (2017) |
| BIRCH | 1 | Dovgan et al. (2019) |
| CFSFDP-HD | 1 | Kausar et al. (2018) |
| CLARANS | 1 | Vasuki and Revathy (2022) |
| DP-means | 1 | Khayi and Rus (2019) |
| Farthest First clustering | 1 | Urbina Nájera et al. (2017) |
| $k$-medians | 1 | C. Wang and Zha (2018) |
| Link-based cluster ensemble | 1 | Iam-On and Boongoen (2017b) |
| Louvain clustering | 1 | Pradana et al. (2020) |
| OPTICS clustering | 1 | Svábensky` et al. (2022) |
| Roche multiway tree | 1 | Q. Yan and Su (2021) |
| $x$-means | 1 | Phanniphong et al. (2019) |